%% file: main.tex
\documentclass[10pt,twocolumn,letterpaper]{article}

\pdfoutput=1

\usepackage{cvpr}

\usepackage{times}
\usepackage{epsfig}
\usepackage{graphicx}
\usepackage{amsmath}

\usepackage{amssymb,amsfonts,mathtools}
\usepackage{algorithm}
\usepackage{algorithmic}
\usepackage{rotating}
\usepackage{bbm}
\usepackage{color}
\usepackage[table]{xcolor}
\usepackage{subcaption}

\usepackage{multicol}
\usepackage{multirow}

\usepackage[breaklinks=true,bookmarks=false]{hyperref}

\newcommand{\Cl}{\mathcal{G}}
\renewcommand{\Re}[1]{\mbox{$\mathbbm{R}^{#1}$}}

\cvprfinalcopy %

\begin{document}

\title{Deep Metric Structured Learning For Facial Expression Recognition}

\author{Pedro D. Marrero Fernandez, Fidel A. Guerrero Pe\~{n}a,\\
Recife Center for Advanced Studies and Systems - CESAR, Brazil\\
{\tt\small \{pdmf, fagp2\}@cesar.org.br}
\and
\: Tsang Ing Ren,\\
Centro de Inform\'atica, Universidade Federal de Pernambuco, Brazil\\
{\tt\small \{tir\}@cin.ufpe.br}
\and
\: Tsang Ing Jyh, \\
University of Antwerp - IMEC, IDLab research group,\\ Sint-Pietersvliet 7, 2000 Antwerp, Belgium\\
{\tt\small \{IngJyh.Tsang\}@uantwerpen.be}
\and
\: Alexandre Cunha\\
Center for Advanced Methods in Biological Image Analysis\\ California Institute of Technology, USA\\
{\tt\small  \{cunha\}@caltech.edu  }}

\maketitle

\input{abstract}

\input{introdution}

\input{method}

\input{experiments}

\input{conclusions}

{\small
\bibliographystyle{ieee_fullname}
\bibliography{refs}
}

\end{document}

%% file: abstract.tex
\begin{abstract}
We propose a deep metric learning model to create embedded sub--spaces with a well defined structure. A new loss function that imposes Gaussian structures on the output space is introduced to create these sub--spaces thus shaping the distribution of the data. Having a mixture of Gaussians solution space is advantageous given its simplified and well established structure. It allows fast discovering of classes within classes and the identification of mean representatives at the centroids of individual classes.
We also propose a new semi--supervised method to create sub--classes. We illustrate our methods on the facial expression recognition problem and validate results on the FER+, AffectNet, Extended Cohn-Kanade (CK+), BU-3DFE, and JAFFE datasets. We experimentally demonstrate that the learned embedding can be successfully used for various applications including expression retrieval and emotion recognition.

\end{abstract}

%% file: introdution.tex
\begin{figure}[t!]
\begin{center}
\includegraphics[width=1.0\linewidth]{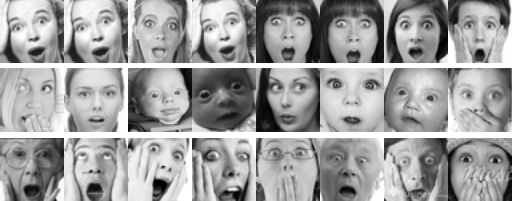}
\end{center}
\vspace{-4mm}
  \caption{{\bf Classes within classes}. The figure depicts some faces in the FER+ dataset classified by our method as having a surprise expression. Our method further separates these faces into other sub--classes, as shown in the three examples above. Each row contains the top eight images identified to be the closest ones to the centroid of their respective sub--class, and each represented by its own Gaussian.  One could tentatively visually describe the top row as faces with strong eye and mouth expressions of surprise, the middle row with mostly mildly surprised eyes, and the bottom row faces with strong surprise expressed with wide open eyes and mouth, and hands on face. Observe the face similarities in each sub--class.}
\label{fig:subset_fer}
{\color{gray}\hrule}
\vspace{-4mm}
\end{figure}

\section{Introduction}

Classical distance metrics like Euclidean distance and cosine similarity are limited and do not always perform well when computing distances between images or their parts. Recently, end--to--end methods \cite{schroff2015facenet, balntas2016learning, song2016deep, wang2014learning} have shown much progress in learning an intrinsic distance metric. They train a network to discriminatively learn embeddings so that similar images are close to each other and images from different classes are far away in the feature space. These methods are shown to outperform others adopting manually crafted features such as SIFT and binary descriptors \cite{dosovitskiy2016discriminative, simo2015discriminative}. Feedforward networks trained by supervised learning can be seen as performing representation learning, where the last layer of the network is typically a linear classifier, {\emph e.g.} a softmax regression classifier.

Representation learning is of great interest as a tool to enable semi-supervised and unsupervised learning. It is often the case that datasets are comprised of vast training data but with relatively little labeled training data. Training with supervised learning techniques on a reduced labeled subset generally results in severe overfitting. Semi-supervised learning is an alternative to resolve the overfitting problem by learning from the vast unlabeled data. Specifically, it is possible to learn good representations for the unlabeled data and use them to solve the supervised learning task.

The adoption of a particular cost function in learning methods imposes constraints on the solution space, whose shape can take any form satisfying the underlying properties induced by the loss function. For example, in the case of triplet loss \cite{schroff2015facenet}, the optimization of the cost function leads to the creation of a solution space where every object has the nearest neighbors within the same class. Unfortunately, it does not generate a much desired probability distribution function, which is achieved by our formulation.

In theory, we would like to have the solution manifold to be a continuous function representing the true original information, because, as in the case of the facial expression recognition problem, face expressions are points in the continuous facial action space resulting from the smooth activation of facial muscles \cite{ekman2002facs}. The transition from one expression to another is represented as the trajectory between the embedded vectors on the manifold surface.

The objective of this work is to offer a formulation for the creation of separable sub--spaces each with a defined structure and with a fixed data distribution. We propose a new loss function that imposes Gaussian structures in the creation of these sub-spaces. In addition, we also propose a new semi-supervised method to create sub--classes within each facial expression class, as exemplified in \figurename~\ref{fig:subset_fer}.

\section{Related Work}

Siamese networks applied to signature verification showed the ability of neural networks to learn compact embedding \cite{bromley1994signature}. OASIS \cite{chechik2010large} and local distance learning \cite{frome2007image} learn fine-grained image similarity ranking models using hand-crafted features that are not based on deep-learning. Recent methods such as \cite{schroff2015facenet, balntas2016learning, song2016deep, wang2014learning} approaches the problem of learning a distance metric by discriminatively training a neural network. Features generated by those approaches are shown to outperform manually crafted features \cite{balntas2016learning}, such as SIFT and various binary descriptors \cite{dosovitskiy2016discriminative, simo2015discriminative}.

Distance Metric Learning (DML) can be broadly divided into contrastive loss based methods, triplet networks, and approaches that go beyond triplets such as quadruplets, or even batch-wise loss. Contrastive embedding is trained on paired data, and it tries to minimize the distance between pairs of examples with the same class label while penalizing examples with different class labels that are closer than a margin $\alpha$ \cite{hadsell2006dimensionality}. Triplet embedding is trained on triplets of data with anchor points, a positive that belongs to the same class, and a negative that belongs to a different class \cite{weinberger2009distance, hoffer2015deep}. Triplet networks use a loss over triplets to push the anchor and positive closer, while penalizing triplets where the distance between the anchor and negative is less than the distance between the anchor and positive, plus a margin $\alpha$. Contrastive embedding has been used for learning visual similarity for products \cite{bell2015learning}, while triplet networks have been used for face verification, person re-identification, patch matching, for learning similarity between images and for fine-grained visual categorization \cite{schroff2015facenet, shi2016embedding, wang2014learning, cui2016fine, balntas2016learning}.

Several works are based on triplet-based loss functions for learning image representations. However, the majority of them use category label-based triplets \cite{zhuang2016fast, wang2017deep, oh2016deep}. Some existing works such as \cite{chechik2010large, wang2014learning} have focused on learning fine-grained representations. In addition, \cite{zhuang2016fast} used a similarity measure computing several existing feature representations to generate ground truth annotations for the triplets, while \cite{wang2014learning} used text image relevance, based on Google image search to annotate the triplets. Unlike those approaches, we use human raters to annotate the triplets. None of those works focus on facial expressions, only recently \cite{vemulapalli2019compact} proposed a system of facial expression recognition based on triplet loss.

%% file: method.tex
\section{Methodology}

\subsection{Structured Gaussian Manifold Loss}

Let $S = \{x_i | x_i\in \Re{D}\}$ be a collection of \textit{ i.i.d.} samples $x_i$ to be classified into $c$ classes, and let $w_j$ represent the $j$--th class, for $j=1,\ldots, c$. The computed class function $l(x) = \arg \max p(w | f_{\Theta}(x))$ returns the class $w_j$ of sample $x$ -- maximum \textit{ a posteriori} probability estimate -- for the neural net function $f_{\Theta}:\Re{D}\to\Re{d}$ drawn independently according to probability $p(x|w_j)$ for input $x$. Suppose we separate $S$ in an embedded space such that each set $C_j = \{x | x \in S, l(x) = w_j\}$ contains the samples belonging to class $w_j$. Our goal is to find a Gaussian representation for each $C_j$ which would allow a clear separation of $S$ in a reduced space, $d \ll D$. 

We assume that $p(f_{\Theta}(x)|w_j)$ has a known parametric form, and it is therefore determined uniquely by the value of a parameter vector $\theta_j$. For example, we might have $p(f_{\Theta}(x)|w_j) \sim N(\mu_j, \Sigma_j)$, where $\theta_j = (\mu_j,\Sigma_j)$, for $N(.,.)$ the normal distribution with mean $\mu_j$ and variance $\Sigma_j$. To show the dependence of $p(f_{\Theta}(x)|w_j)$ on $\theta_j$ explicitly, we write $p(f_{\Theta}(x)|w_j)$ as $p(f_{\Theta}(x)|w_j,\theta_j)$. Our problem is to use the information provided by the training samples to obtain a good transformation function $f_{\Theta}(x_j)$ that generates embedded spaces with a known distribution associated with each category. Then the \textit{ a posteriori} probability $P(w_j|f_{\Theta}(x))$ can be computed from $p(f_{\Theta}(x)|w_j)$ by the Bayes' formula:

\begin{equation}
P(w_j| f_{\Theta}(x) ) = \frac{p(w_j)p(f_{\Theta}(x)|w_j, \theta_i)}{ \sum_i^c p(w_i)p(f_{\Theta}(x)|w_i, \theta_i) }
\end{equation}

We use the normal density function for $p(x|w_j,\theta_j)$. The objective is to generate embedded sub-spaces with defined structure. Thus, using the Gaussian structures:

\begin{equation}                                      
p(f_{\Theta}(x)|w_j,\mu_j,\Sigma_j) = \frac{1}{(2\pi)^{n/2}|\Sigma_j|^{1/2} } \exp( -\frac{1}{2} X^T \Sigma_j^{-1} X )    
\end{equation}
where $X=(f_{\Theta}(x)-\mu_j)$. For the case $\Sigma_j = \sigma^2I$, where $I$ is the identity matrix:
\begin{equation}
p(x|w_j,\mu_j,\sigma_j) = \frac{1}{ \sqrt[]{(2\pi)^n} \sigma_j } \exp( -\frac{||f_{\Theta}(x)-\mu_j||^2}{2\sigma_j^2} )   
\end{equation}

In a supervised problem, we know the \textit{a posteriori} probability $P(w_j|x)$ for the input set. From this, we can define our structured loss function as the mean square error between the \textit{a posteriori} probability of the input set and the \textit{a posteriori} probability estimated for the embedded space:

\begin{equation} \label{eq:gd}
\mathcal{L}_{rep} = \mathbb{E} \left \{ ||P(w_j|f_{\Theta}(x_i)) - P(w_j|x_i)||^2_2 \right \} 
\end{equation}

We applied the steps described in Algorithm \ref{alg:sgm} to train the system. The batch size is given by $n \times c$ where $c$ is the number of classes, and $n$ is the sample size. In this work, we use  $n = 30$, thus for eight classes the batch size is 240, which was used for the estimation of the parameters in Equation \ref{eq:gd}.

\begin{algorithm}[h]
\caption{Structured Gaussian Manifold Learning. $f_{\Theta}$: Neural Network; $S$: dataset; $C_j$ are the subset of the elements of class $w_j$; $N$: number of updates;  }
\label{alg:sgm}
\begin{algorithmic}[1]

\STATE $k \gets 0 $
\WHILE{ $k<N$ }
\STATE \{Sample$(x_i,w_i)\} \sim S $, get current batch.  
\STATE $z \gets f_{\Theta}(x_i),  $ representation. 
\STATE $\theta_j \gets \{ \mu_{j}, \sigma{j} \}$ \\ where $\sigma$ is a parameters ($\sigma=0.5$ in this work) and $\mu_j$ is the mean of the elements of the class $w_j$:  \\
$$ \mu_j = \frac{1}{ |C_j| }\sum_{k \in C} z_k $$ 

where $|.|$ denotes set cardinality.

\STATE Evaluation of the Loss function. For the explanation of the loss representation see equation \ref{eq:gd}: 

$$\mathcal{L} \gets \mathcal{L}_{rep}(z_i, w_i, \mu, \sigma ) + \frac{1}{|\Omega|} \sum_k||f_{\Theta}(x_k)||_2 $$
where $x_k \in \Omega$
\STATE $\Theta^{t+1} = \Theta^t - \nabla \mathcal{L}$, backward and optimization steps.
\ENDWHILE

\end{algorithmic}
\end{algorithm}

We define the accuracy of the model as the ability of the parameter vector $\theta$ to represent the test dataset in the embedded space.

\subsection{Deep Gaussian Mixture Sub-space}

The same facial expression may possess a different set of global features. For example, ethnicity can determine specific color and shape, while age provides physiological differences of facial characteristics; moreover, gender, weight, and other features can determine different facial characteristics, while having the same expression. Our proposal can group and extract these characteristics automatically. We propose to represent each facial expression class as a Gaussians Mixture. These Gaussian parameters are obtained in an unsupervised way as part of the learning processes. We start from a representation space given by Algorithm \ref{alg:sgm}. Subsequently, a clustering algorithm is applied to separate each class into a new class subset. This process is repeated until reaching the desired granularity level. Algorithm \ref{alg:msgm} shows the set of steps to obtain the new sub-classes.

\begin{algorithm}[h]
\caption{Deep Gaussian Mixture Sub-space Learning. $L$: Maximum level of subdivisions for the class; $f_{\Theta}$: Neural Network; $StructureGaussianManifold$: Structure Gaussian Manifold Algorithm \ref{alg:sgm}; $EM$: Expectation Maximization Algorithm; $S$: dataset; $C_j$ are the subset of the elements of class $w_j$; $N$: number of updates;}

\label{alg:msgm}
\begin{algorithmic}[1]

\STATE $l \gets 1 $
\STATE $X,Y \sim S $
\STATE $\hat{Y} \gets Y$
\WHILE{ $l<L$ }
\STATE $\Theta \gets$StructureGaussianManifold$( \{X, \hat{Y} \}, N )$
\STATE $Z = f_{\Theta}(X) $ 
\STATE $\hat{Y} \gets  \{ \oslash \} $
\STATE $k \gets 0$
\FORALL{ class $w_j$ }
\STATE $Z_c = \{ z ~ | ~ \forall z \in C_j  \}$
\STATE $g \gets EM(l, Z_c )$
\STATE $\hat{Y} =  \{ \hat{Y}, g+k  \}$
\STATE $k+=l$
\ENDFOR
\ENDWHILE

\end{algorithmic}
\end{algorithm}

%% file: experiments.tex
\section{Experiments}

\subsection{Protocol}

For the evaluation of the clustering task, we use the F1-measure and Normalized Mutual Information (NMI) measures. The F1-measure computes the harmonic mean of the precision and recall, $F1 = \frac{2PR}{P+R}$. The NMI measure take as input a set of clusters $\Omega = \{o_1, \ldots, o_k\}$ and a set of ground truth classes $\Cl = \{g_1, \ldots, g_k\}$, $o_i$ indicates the set of examples with cluster assignment $i$ and $g_j$ indicates the set of examples with the ground truth class label $j$. Normalized mutual information is defined by the ratio of mutual information and the average entropy of the clusters and the entropy of the labels, $NMI(\Omega, \Cl) = \frac{I(\Omega;\Cl)}{2(H(\Omega)+H(\Cl))}$, for complete details see \cite{manning2008introduction}. For the retrieval task, we use the Recall@K \cite{jegou2011product} measure. Each test image (query) first retrieves K Nearest Neighbour (KNN) from the test set and receives score $1$ if an image of the same class is retrieved among the KNN, and $0$ otherwise. Recall@K averages those score over all the images. Moreover, we also evaluate accuracy, i.e. the fraction of results that are the same class as queried image, averaged over all queries. While the classification task is evaluated using KNN.

For the training process, we use the Adam method \cite{adamDBLP} with a learning rate of 0.0001 and batch size of 256 (samples of size 32 to estimate the parameters in each iteration). In the TripletLoss case, we used 128 triplets in each batch. The neural networks were initialized with the same weights in all cases.

\subsection{Result}
\subsubsection{Representation and Recover}

The groups used for the evaluation of the measures are obtained using $K$-means, whereas $K$ equals the number of classes (8 in the case of the FER+ \cite{BarsoumICMI2016}, AffectNet \cite{picard2000affective}, CK+ \cite{lucey2010extended} datasets, and 7 for JAFFE \cite{lyons1998coding} and BU-3DFE \cite{yin20063d} datasets).

The results obtained for the clustering task show that the proposed method presents good group quality (see table \ref{tab:metricNMIforall}) in similar domains. As can be observed, the results are degraded for different domains. In general, we observe that the TripletLoss is most robust to the change of domains on all models. However, the best result is achieved using the proposed method for the RestNet18 \cite{resnetDBLP} model in FER+, CK+, and BU-3DFE.

\begin{table*}[!t]
\begin{center}
\begin{tabular}{|c|c|c|c|c|c|c|}

\hline
Method & Arch. & FER+$^\dag$ & AffectNet$^\ddag$ & JAFFE & CK+ & BU-3DFE \\
\hline\hline

\multirow{5}{*}{TripletLoss}    &FMPNet\cite{fmpYuISFERMDEEP}           & 55.257 & 10.627 & 19.528 & 71.129 & 34.901   \\
                                &CVGG13\cite{FerpBarsoumICMI2016}          & 67.384 &  9.103 & 28.295 & 68.303 & 27.275   \\
                                &AlexNet\cite{alexnetDBLP}         & 67.035 & 12.945 & 30.241 & 68.800 & 27.039   \\
                                &ResNet18\cite{resnetDBLP}         & 64.457 & \textbf{15.588} & \textbf{31.046} & 74.028 & 36.708   \\
                                &PreActResNet18\cite{he2016identity}  & 57.904 &  8.452 & 20.699 & 70.079 & 27.580   \\

\hline\hline

\multirow{5}{*}{SGMLoss}        &FMPNet\cite{fmpYuISFERMDEEP}           & 57.880 & 10.469 & 26.196 & \textbf{77.839} & 36.559   \\
                                &CVGG13\cite{FerpBarsoumICMI2016}           & 65.139 & 10.355 & 24.293 & 66.062 & 27.233   \\
                                &AlexNet\cite{alexnetDBLP}          & 62.091 & 10.582 & 24.560 & 65.230 & 28.115   \\
                                &ResNet18\cite{resnetDBLP}          & \textbf{68.840} & 12.333 & 30.382 & \textbf{77.902} & \textbf{37.545}   \\
                                &PreActResNet18\cite{he2016identity}  & 51.425 &  6.886 & 23.216 & 61.413 & 26.104   \\

\hline
\end{tabular}
\end{center}\vspace{-3mm}
\caption{The NMI (\%) of the clustering task for all datasets applying the TripletLoss and SGMLoss models trained on FER+. SGMLoss: Structured Gaussian Manifold Loss, Arch: Architecture, FER+$^\dag$: FER+ test dataset, AffectNet$^\ddag$: validation dataset. JAFFE CK+ and BU3DFE datasets are used as test set.}
\label{tab:metricNMIforall}

{\color{lightgray}\hrule}\vspace{-2mm}
\end{table*}

\begin{figure}
\centering
\includegraphics[width=0.8\linewidth]{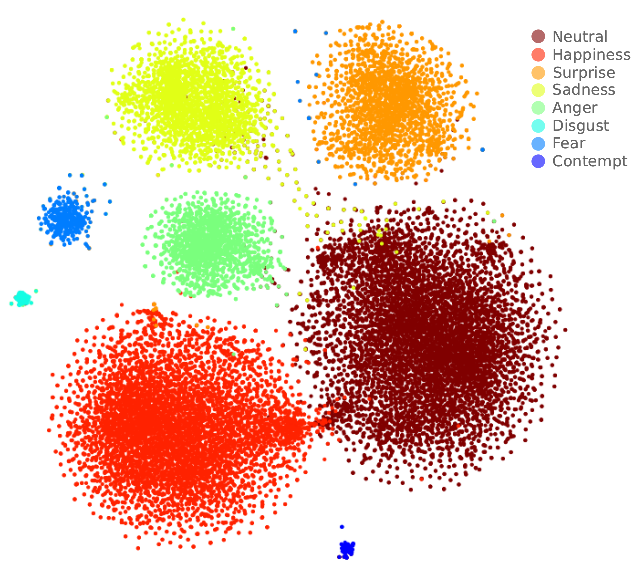}
\caption{Barnes-Hut t-SNE visualization \cite{van2014accelerating} of the SGMLoss for the FER+ database. Each color represents one of the eight emotions including neutral. }
\label{fig:embedded}
{\color{lightgray}\hrule}\vspace{-2mm}
\end{figure}

\figurename~\ref{fig:embedded} shows a 2D t-SNE \cite{van2014accelerating} visualization of the learned SGMLoss embedding space using the FER+ training set. The amount of overlap between the two categories in this figure roughly demonstrates the extent of the visual similarity between them. For example, happy and neutral have some objects overlap, indicating that these cases could be confused easily, and both of them have a very low overlap with fear indicating that they are visually very distinct from fear. Also, the spread of a category in this figure indicates the visual diversity within that category. For example, happiness category maps to some distinct regions indicating that there are some visually distinct modes within this category.

\begin{figure}
\centering
\includegraphics[width=0.8\linewidth]{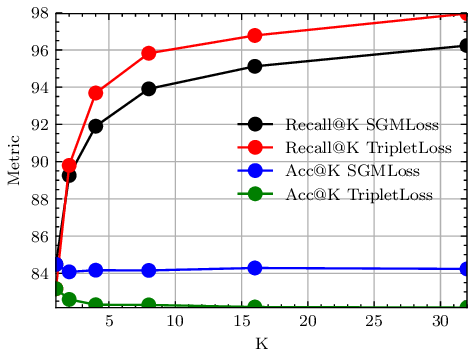}
\caption{Recall@K and Acc@K measures for the test split FER+ dataset. The applied model was the ResNet18 having $K=\{1, 2, 4, 8, 16, 32\}$.}
\label{fig:ferprecall}
{\color{lightgray}\hrule}\vspace{-2mm}
\end{figure}

\begin{figure*}
\centering
\includegraphics[width=0.8\linewidth]{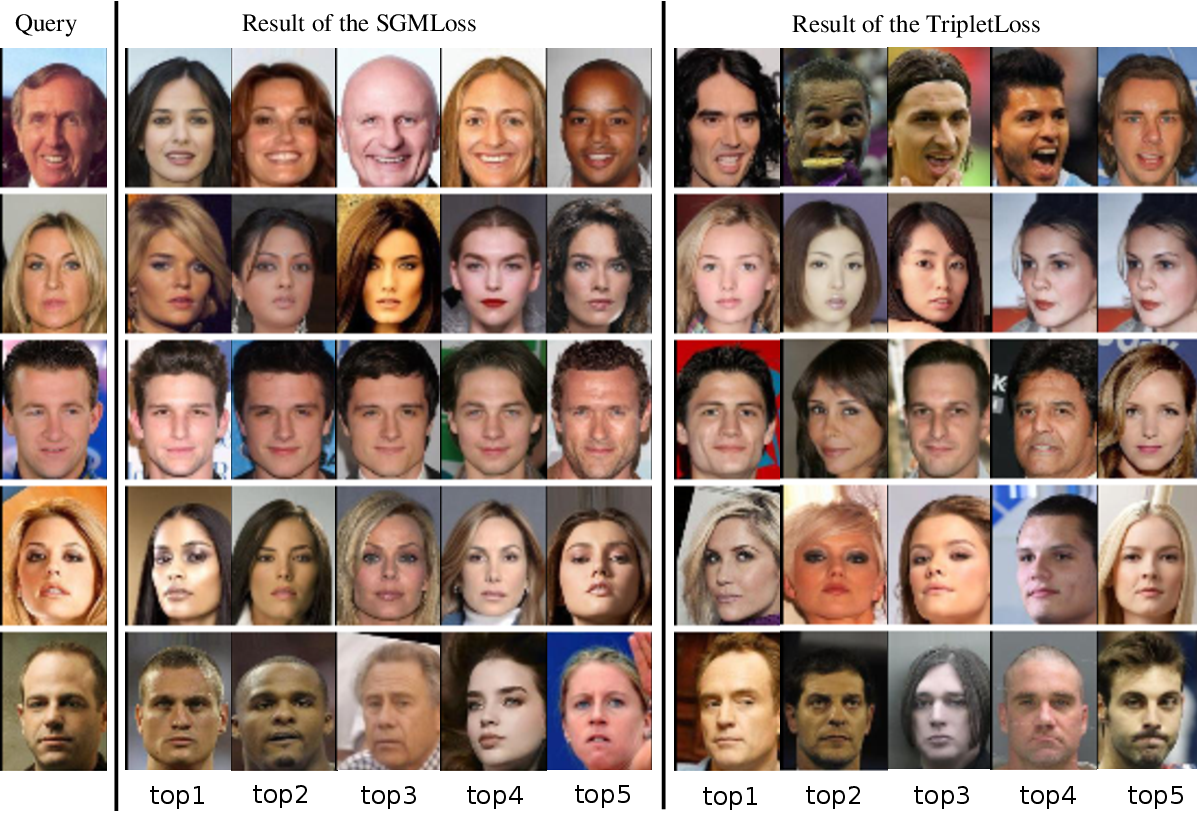}
\caption{Top-5 images retrieved using SGMLoss (left) and TripletLoss (right) embeddings. The overall results of the SGMLoss match the query set apparently better when compared to TripletLoss.}
\label{fig:recovery_celeba}
{\color{lightgray}\hrule}\vspace{-2mm}
\end{figure*}

\figurename~\ref{fig:ferprecall} shows the results obtained in the recovery task (Recall@K and Acc@K measures) for $K=\{1,2,4,8,16,32\}$. TripletLoss obtains better recovery results for all K but to the detriment of accuracy. Our method manages to increase its recovery value while preserving quality. It means that most neighbors are of the same class. \figurename~\ref{fig:recovery_celeba} shows the top-5 retrieved images for some of the queries on CelebA dataset \cite{liu2015deep}. The overall results of the proposed SGMLoss embedding are clearly better than the results of TripletLoss embedding.

\subsubsection{Classification}

\begin{figure}
\centering
\includegraphics[width=0.9\linewidth]{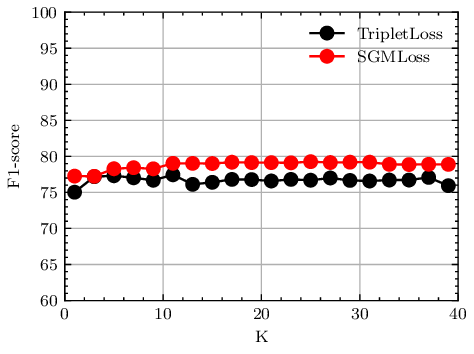}
\caption{Classification performance of the SGMLoss and TripletLoss on the FER+ validation set when combined with KNN classifier.  }
\label{fig:crossval}
{\color{lightgray}\hrule}\vspace{-2mm}
\end{figure}

The proposed SGMLoss method can be used for FER by combining it with the KNN classifier. \figurename~\ref{fig:crossval} shows the average F1-score of the SGMLoss and TripletLoss on the FER+ validation set as a function of the number of neighbors used. F1-score is maximized for K=11.

Table \ref{tab:metricF1all} compares the classification performance of the SGMLoss embedding (using 11 neighbors) with TripletLoss and CNN models. In general, our method obtains the best classification results for all architectures. ResNet18 CNN model does not obtains a significant higher accuracy.  Moreover, our results surpass the accuracy 84.99 presented in \cite{BarsoumICMI2016}.

\begin{table*}[!t]

\begin{center}
\begin{tabular}{ |c|c|c|c|c|c|c| }
\hline
            Method              & Arch.         & Acc. & Prec. & Rec. & F1             \\

\hline\hline

\multirow{5}{*}{CNN}            &FMPNet\cite{fmpYuISFERMDEEP}              & 79.535 & 66.697 & \textbf{68.582} & 67.627           \\
                                &CVGG13\cite{FerpBarsoumICMI2016}          & 84.316 & 75.151 & 67.425 & 71.079                    \\
                                &AlexNet\cite{alexnetDBLP}                 & 86.038 & 77.658 & \textbf{68.657} & 72.881           \\
                                &ResNet18\cite{resnetDBLP}                 & \textbf{\underline{87.695}} & 85.956 & \textbf{\underline{69.659}} & 76.954  \\
                                &PreActResNet18\cite{he2016identity}       & 82.372 & 76.915 & 65.238 & 70.597                    \\

\hline\hline

\multirow{5}{*}{TripletLoss}    &FMPNet\cite{fmpYuISFERMDEEP}              & 82.563 & \textbf{79.554} & 62.406 & 69.944   \\
                                &CVGG13\cite{FerpBarsoumICMI2016}          & 85.974 & 82.034 & \textbf{68.112} & 74.428   \\
                                &AlexNet\cite{alexnetDBLP}                 & 86.038 & 80.598 & 67.895 & 73.703            \\
                                &ResNet18\cite{resnetDBLP}                 & 87.121 & 78.543 & 68.378 & 73.109            \\
                                &PreActResNet18\cite{he2016identity}       & 83.519 & 74.081 & 64.856 & 69.162            \\

\hline\hline

\multirow{5}{*}{SGMLoss}        &FMPNet\cite{fmpYuISFERMDEEP}              & \textbf{83.360} & 78.806 & 66.520 & \textbf{72.143}            \\
                                &CVGG13\cite{FerpBarsoumICMI2016}          & \textbf{86.261} & \textbf{86.321} & 67.341 & \textbf{75.659}   \\
                                &AlexNet\cite{alexnetDBLP}                 & \textbf{86.643} & \textbf{86.182} & 67.673 & \textbf{75.814}   \\
                                &ResNet18\cite{resnetDBLP}                 & 87.631 & \textbf{88.614} & 68.724 & \textbf{\underline{77.412}}            \\
                                &PreActResNet18\cite{he2016identity}       & \textbf{84.316} & \textbf{\underline{89.008}} & \textbf{66.519} & \textbf{76.138}   \\

\hline
\end{tabular}
\end{center}\vspace{-2mm}
\caption{Classification results of the CNN, TripletLoss and SGMLoss models trained on FER+. SGMLoss: Structured Gaussian Manifold Loss, Arch: Architecture, FER+$^\dag$: FER+ test dataset, AffectNet$^\ddag$ validation dataset.}
\label{tab:metricF1all}

{\color{lightgray}\hrule}\vspace{-2mm}
\end{table*}

\begin{figure}
\centering
\includegraphics[width=0.8\linewidth]{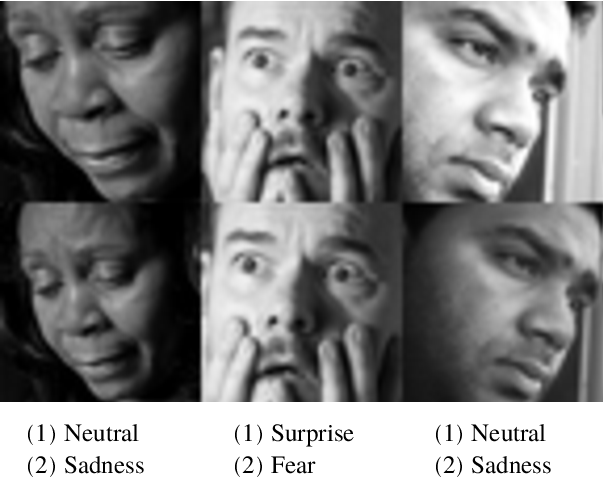}
\caption{Examples of mislabeled images on the FER+ dataset that were recovery using SGMLoss. The first row show the result of the query (1) and the second row the result (2). We can clearly observer that two very similar images have different labels in the dataset.}  %
\label{fig:neighbors}
{\color{lightgray}\hrule}\vspace{-4mm}
\end{figure}

The Facial Expression dataset constitute a great challenge due to the subjectivity of the emotions \cite{Marrero-Fernandez2014EvaluatingRecognition}. The labeling process requires the effort of a group of specialists to make the annotations. FER+ and AffectNet datasets contains many problems in the labels. In \cite{BarsoumICMI2016} an effort was made to improve the quality of the labels of the FER+ (dataset used in our experiments) by re-tagging the dataset using crowd sourcing. \figurename~\ref{fig:neighbors} shows some mislabeled images retrieved by our method. The scale, position, and context could influence the decision of a non-expert tagger such as those in crowd sourcing.

Experimental results show the quality of the embedded representation obtained by SGMLoss in the classification problems. Our representation improves the representation obtained by TripletLoss, which is the method most used in the identification and representation problems.

\subsubsection{Clustering}

For the training process, we use the Adam method \cite{adamDBLP} with a learning rate of 0.0001, a batch size of 640 and 500 epoch. The maximum level of subdivision used is L=5 (this value guarantee that the batch for a subclass in this level to be 128). The ResNet18 architecture is selected to train the FER+ dataset. The objective of this experiment is to visually analyse the clustering obtained by this approach.

\begin{figure}
\centering
\includegraphics[width=0.8\linewidth]{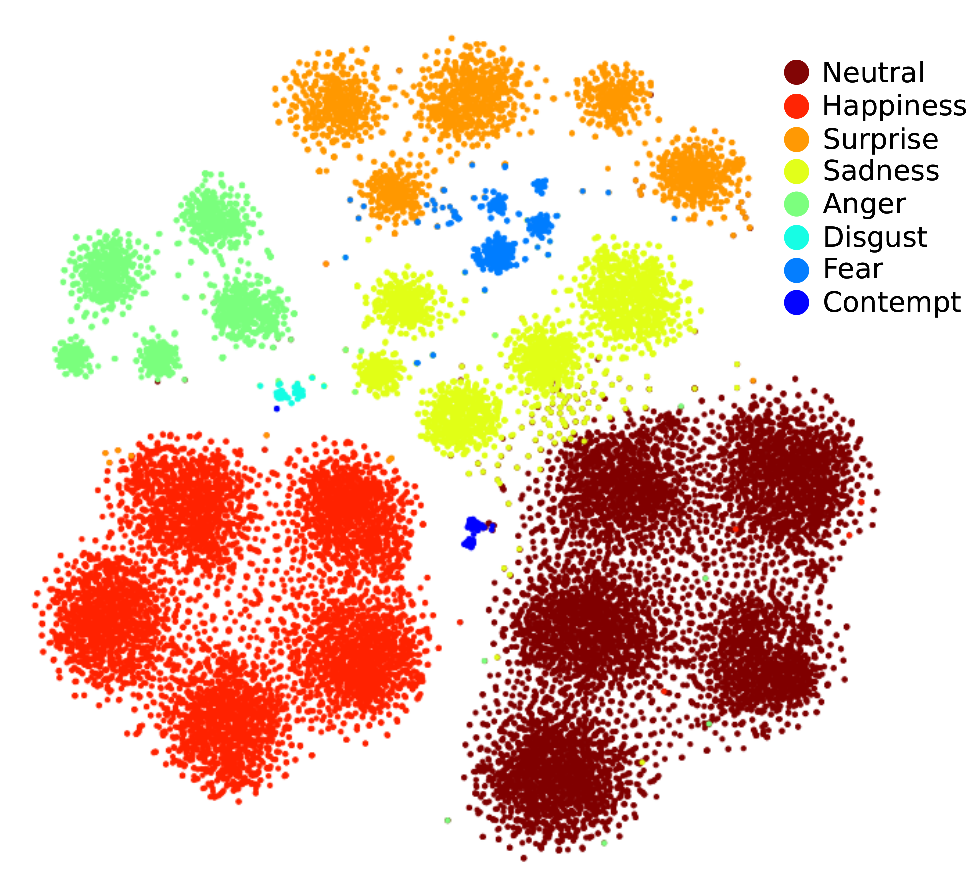}
\caption{Barnes-Hut t-SNE visualization \cite{van2014accelerating} of the Deep Gaussian Mixture Sub-space for the FER+ database. Each color represents one of the eight emotions including neutral. }
\label{fig:mitosis}
{\color{lightgray}\hrule}\vspace{-2mm}
\end{figure}

The results shown in \figurename~\ref{fig:mitosis} present 64-dimensional embedded space using the Barnes-Hut t-SNE visualization scheme \cite{van2014accelerating} using the Deep Gaussian Mixture Sub-space model for the FER+ dataset. The method created five Gaussian sub-spaces for the unsupervised case for each class.

For the clustering task, all embedded vectors are calculated and EM method is applied creating 40 groups. For each group, the medoid is calculated. The medoid is the object in the group closest to the centroid (mean to the sample). The Top-k of a group contains the \textit{k-objects} nearer to the medoid of the group.

\begin{figure}
\centering
\begin{tabular}{c}
\includegraphics[width=0.9\linewidth]{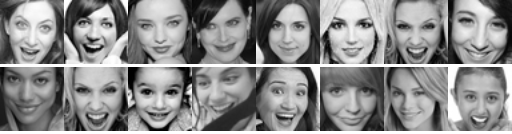}   \\
(a)\\
\includegraphics[width=0.9\linewidth]{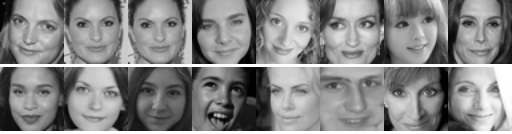}   \\
(b)\\
\includegraphics[width=0.9\linewidth]{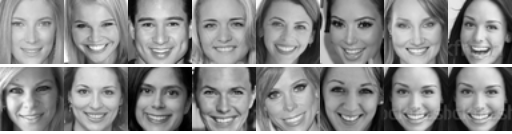}   \\
(c)\\
\includegraphics[width=0.9\linewidth]{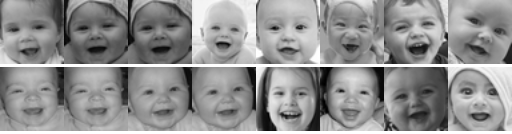}   \\
(d)\\
\includegraphics[width=0.9\linewidth]{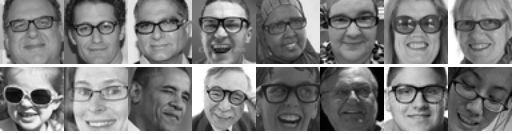}   \\
(f)\\
\end{tabular}
\caption{Top-16 images of the clustering obtained from the class happiness on the FER+ dataset. }
\label{fig:ferp_grup}
{\color{lightgray}\hrule}\vspace{-2mm}
\end{figure}

\figurename~\ref{fig:ferp_grup} shows the Top-16 images obtained for the happiness category.  The first group (\figurename~\ref{fig:ferp_grup} (a) ) shows an expression of happiness closer to surprise (raised eyebrows and open mouth) with the shape of the eyes similar to each other. The second group (\figurename~\ref{fig:ferp_grup} (b)) represents an expression closer to contempt.  The third group (\figurename~\ref{fig:ferp_grup} (c)) shows an expression of more intense happiness (the teeth are shown in all cases) with the shape of the mouth very similar to each other. In the fourth case (\figurename~\ref{fig:ferp_grup}  (d) ) shows a subcategory that is present in all facial expressions. Babies are a typically expected subset due to the intensity of expression and the physiognomical formation. Generally babies and children from 1 to 4 years old present facial expressions of greater intensity. The last group (\figurename~\ref{fig:ferp_grup} (f) ) represents people with glasses and large eyes.

\begin{figure}
\centering
\begin{tabular}{c}
\includegraphics[width=0.9\linewidth]{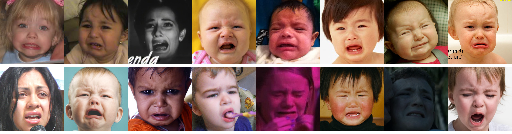}  \\
(a)\\
\includegraphics[width=0.9\linewidth]{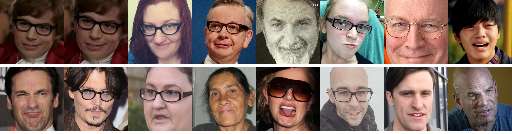}  \\
(b)\\
\includegraphics[width=0.9\linewidth]{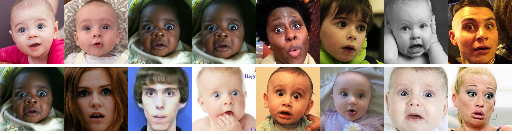}   \\
(c)\\
\includegraphics[width=0.9\linewidth]{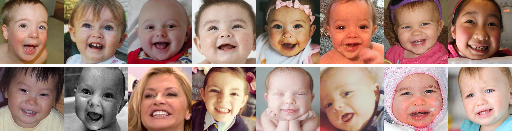}  \\
(d)\\
\includegraphics[width=0.9\linewidth]{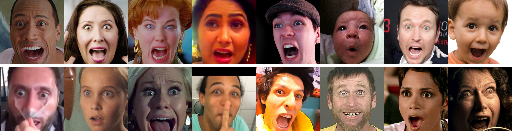}  \\
(f)\\
\end{tabular}
\caption{ Top-16 images of the 5 clustering obtained from the AffectNet dataset. }
\label{fig:affect_grup}
{\color{lightgray}\hrule}\vspace{-4mm}
\end{figure}

The presented method is a powerful tool for tasks such as photo album summarization. In this task, we are interested in summarizing the diverse expression content present in a given photo album using a fixed number of images. \figurename~\ref{fig:affect_grup} shows 5 of the 40 groups obtained on AffectNet dataset. The obtained groups show great similarity in terms of FER. These results demonstrate the generalization capacity of the proposed method and its applicability to problems of FER clustering.

%% file: conclusions.tex
\section{Conclusions}

We introduced two new metric learning representation models in this work, namely Deep Gaussian Mixture Subspace Learning and Structured Gaussian Manifold Learning. In the first model, we build a Gaussian representation of expressions leading to a robust classification and grouping of facial expressions. We illustrate through many examples, the high quality of the vectors obtained in recovery tasks, thus demonstrating the effectiveness of the proposed representation. In the second case, we provide a semi-supervised method for grouping facial expressions. We were able to obtain embedded subgroups sharing the same facial expression group. These subgroups emerged due to shared specific characteristics other than the general appearance. For example, individuals with glasses expressing a happy appearance.